\documentclass[10pt,twocolumn,letterpaper]{article}

\usepackage{wacv}
\usepackage{times}
\usepackage{epsfig}
\usepackage{graphicx}
\usepackage{amsmath}
\usepackage{amssymb}

\wacvfinalcopy 

\def\wacvPaperID{****} 

\ifwacvfinal\fi
\begin{document}

\title{Crowd Counting Using Scale-Aware Attention Networks}


\author{Mohammad Asiful Hossain \hspace{1.2cm}  Mehrdad Hosseinzadeh \hspace{0.7cm} Omit Chanda \hspace{0.7cm} Yang Wang\\
University of Manitoba\\
{\tt\small \{hossaima, mehrdad, omitum, ywang\}@cs.umanitoba.ca}
}


\maketitle
\ifwacvfinal\thispagestyle{empty}\fi

\begin{abstract}
In this paper, we consider the problem of crowd counting in images. Given an image of a crowded scene, our goal is to estimate the density map of this image, where each pixel value in the density map corresponds to the crowd density at the corresponding location in the image. Given the estimated density map, the final crowd count can be obtained by summing over all values in the density map. One challenge of crowd counting is the scale variation in images. In this work, we propose a novel scale-aware attention network to address this challenge. Using the attention mechanism popular in recent deep learning architectures, our model can automatically focus on certain global and local scales appropriate for the image. By combining these global and local scale attentions, our model outperforms other state-of-the-art methods for crowd counting on several benchmark datasets.

\end{abstract}


\section{Introduction}\label{sec:intro}

We consider the problem of crowd counting in arbitrary static images. Given an arbitrary image of a crowded scene without any prior knowledge about the scene (e.g. camera position, scene layout, crowd density), our goal is to estimate the density map of the input image, where each pixel value in the density map corresponds to the crowd density at the corresponding location of the input image. The crowd count can be obtained by integrating the entire density map. In particular, we focus on the setting where the training data have dotted annotations, i.e. each object instance (e.g. people) is annotated with a single point in the image. 

Crowd counting has many real-world applications, such as surveillance, public safety, traffic monitoring, urban planning~\cite{sam17_cvpr}. The methods developed for crowd counting can also be used in counting objects in many other domains, such as counting cells or bacteria in microscopic images \cite{zhangY16_cvpr}, counting animals for ecologic studies \cite{arteta16_eccv}, counting vehicles in traffic control~\cite{rubio16_eccv,zhang17_cvpr,zhangS17_iccv}.

The challenges of crowd counting are manifold, including severe occlusion, perspective distortion, diverse crowd densities, and so on (see Fig.~\ref{fig:sample} for some sample images). Some early work on crowd counting is based on head detection~\cite{idrees13_cvpr}. In recent years, convolutional neural networks (CNNs) have become popular in crowd counting. Most of the CNN-based approaches~\cite{zhangC15_cvpr,zhangY16_cvpr,sam17_cvpr,sindagi17_iccv} work by estimating a density map from the image, then obtain the crowd count based on the density map. The accuracy of crowd counting largely depends on the quality of the estimated density map.

\begin{figure}[t]
  \begin{center}
  {\setlength\tabcolsep{1pt}
  \begin{tabular}{cccc}
    \includegraphics[width=1.1in,height=.98in]{qualitative_images/final_images/IMG_6.jpg}&
    \includegraphics[width=1.1in,height=.98in]{qualitative_images/final_images/IMG_12.jpg}&
    \includegraphics[width=1.1in,height=.98in]{qualitative_images/final_images/IMG_316.jpg}\\

  \end{tabular}
  }
  
  \end{center}
  \caption{Sample images from crowd counting datasets. These images tend to have severe occlusions, perspective distortion, diverse crowd densities, etc. These factors make crowd counting a challenging problem.}
  \label{fig:sample}
\end{figure}

Due to the variations of people/head sizes in crowded scenes, a standard feed-forward CNN model with a single scale usually cannot effectively capture the diverse crowd densities in an image. Several recent work~\cite{zhangY16_cvpr,sam17_cvpr,sindagi17_iccv} has demonstrated the importance of using multi-scale CNN architectures in crowd counting. Zhang et al.~\cite{zhangY16_cvpr} propose a multi-column CNN (MC-CNN) architecture with several branches, where each branch uses filters of different sizes. The features from these branches are combined together for the density map estimation. The final features will capture the multi-scale information of the image. Sam et al.~\cite{sam17_cvpr} use a similar intuition and develop an architecture called switching convolutional neural network (Switch-CNN). Switch-CNN also uses multiple branches to extract features using filters of different sizes. However, instead of concatenating the features maps from all the branches, Switch-CNN learns a classifier that predicts a discrete density class (i.e. scale) of an input image, then uses this predicted scale to choose one of the branches and uses the features from that branch for the density estimation. Sindagi et al.~\cite{sindagi17_iccv} propose the contextual pyramid CNNs (CP-CNN). In addition to global density class of the whole image, CP-CNN also predicts the local density class for each patch in the image. The contextual information obtained from both global and local density class predictions are combined for the final density estimation. 

In recent years, attention models have shown great success in various computer vision tasks~\cite{mnih14_nips,xu15_icml}. Instead of extracting features from the entire image, the attention mechanism allows models to focus on the most relevant features as needed. Our model is partly inspired by some work of Chen et al.~\cite{chen16_cvpr} in semantic segmentation. This work uses attentions to focus on the relevant scale dimension of an image for semantic segmentation. In this paper, we use a similar idea and develop a CNN architecture with scale-aware attentions for crowd density estimation and counting. The attention in our model plays a similar role to the ``switch'' (i.e. density classifier) in Switch-CNN~\cite{sam17_cvpr}. Switch-CNN makes a hard decision by selecting a particular scale based on the density classifier output and only uses the features corresponding to that scale for the final prediction. The problem is that if the density classifier is not completely accurate, it might select the wrong scale and lead to incorrect density estimation in the end. In contrast, the attention is our model acts as a ``soft switch''. Instead of selecting a particular scale, we re-weight the features of a particular scale based on the attention score corresponding to that scale.

The contributions of this paper are manifold. First, we introduce the attention mechanism in CNN-based crowd counting models. Although attention models have been successful in many other vision tasks, our work is the first to use attention models in crowd counting. Second, previous work usually uses attention models to focus on certain spatial locations in an image. In our work, we instead use attentions to focus on certain scales. Previous methods~\cite{zhangY16_cvpr,sam17_cvpr,sindagi17_iccv} in crowd counting select the scale using a learned classifier. In contrast, the attention mechanism allows our model to ``softly'' select the scales. Finally, we demonstrate that our proposed approach outperforms other state-of-the-art methods on several benchmark datasets.

\section{Related Work}\label{sec:related}
Most existing crowd counting approaches work by first extracting low-level features from images, then map these features to density maps or counts using various techniques. Loy et al. \cite{loy} has categorized the existing methods into three groups: (1) detection-based methods (2) regression-based methods, and (3) density estimation based methods. In the following, we briefly review some of these methods. Interested reader may refer to \cite{loy} for a more extensive review.

Early work~\cite{detection} on crowd counting uses detection-based approaches. These approaches usually apply a person or head detector on an image. Detection-based approaches often cannot handle high density crowd. To address the limitation of detection-based methods, some work~\cite{idrees13_cvpr} uses a regression-base method that directly learns the mapping from an image patch to the count.

In recent years, convolutional neural networks (CNNs) have been popular in almost all vision tasks including crowd counting. Walach et al.~\cite{walachwcnn} propose a method for learning CNN-based crowd counting model in a layer-wise fashion. Zhang et al.~\cite{zhangC15_cvpr} introduce a cross-scene crowd counting method by fine-tuning a CNN model to the target scene.

One particular challenge in crowd counting is the scale variation in crowd images. Zhang et al. \cite{zhangY16_cvpr} propose a multi-column architecture (MC-CNN) for crowd counting. This multi-column architecture uses three branches of CNNs. Each branch works at a different scale level of the input image. The three branches are fused in the end to produce the output. Onoro-Rubio and L\'opez-Sastre \cite{rubio16_eccv} addressed the scale variation issue by proposing a scale--aware counting model called Hydra CNN. This model is formulated as a regression model in which the network
learns  the way of mapping of  the  image  patches  to  their  corresponding  object  density  maps. Also, Boominathan et al. \cite{boominathan16_multimedia} tackle the issue of scale variation using a combination of deep and shallow networks. 

Although these methods are proved to be robust to scale variations, they have some adverse effect that limits the size of input image during training. By reducing the training image size, they would not be capable of learning the features of original image size. To address this drawback, Sindagi et al. \cite{sindagi17_avss} presented a novel end--to--end cascaded 
CNN that jointly produces the estimated count and high quality density map. The high--level features of their network enables it to learn globally relevant and discriminative features. 

 Sam et al. \cite{sam17_cvpr} introduce a coarse to fine Switch-CNN network that chooses a branch corresponding to an estimated scale instead of fusing all branches. Sindagi et al. \cite{sindagi17_iccv} develop a contextual pyramid CNN (CP-CNN) that combines both global and local contextual information for crowd counting. Most recently, Liu et al.~\cite{decidenet} introduce a method called DecideNet. DecideNet has two counting models -- a regression mode and a detection mode. Depending on the real density condition at a location in an image, DecideNet will learn to automatically switch between these two modes.

Inspired by the previous work, we propose a new approach for handling scale variations in crowd counting. Our proposed model uses attentions to automatically focus on a particular scale, both at the whole image level and at the local patch level. Our model is conceptually simpler than DecideNet~\cite{decidenet}, since we do not need to switch between two different counting modes. 

\begin{figure*}[t]
\centering
\includegraphics[width=6.7in, height=3.3in]{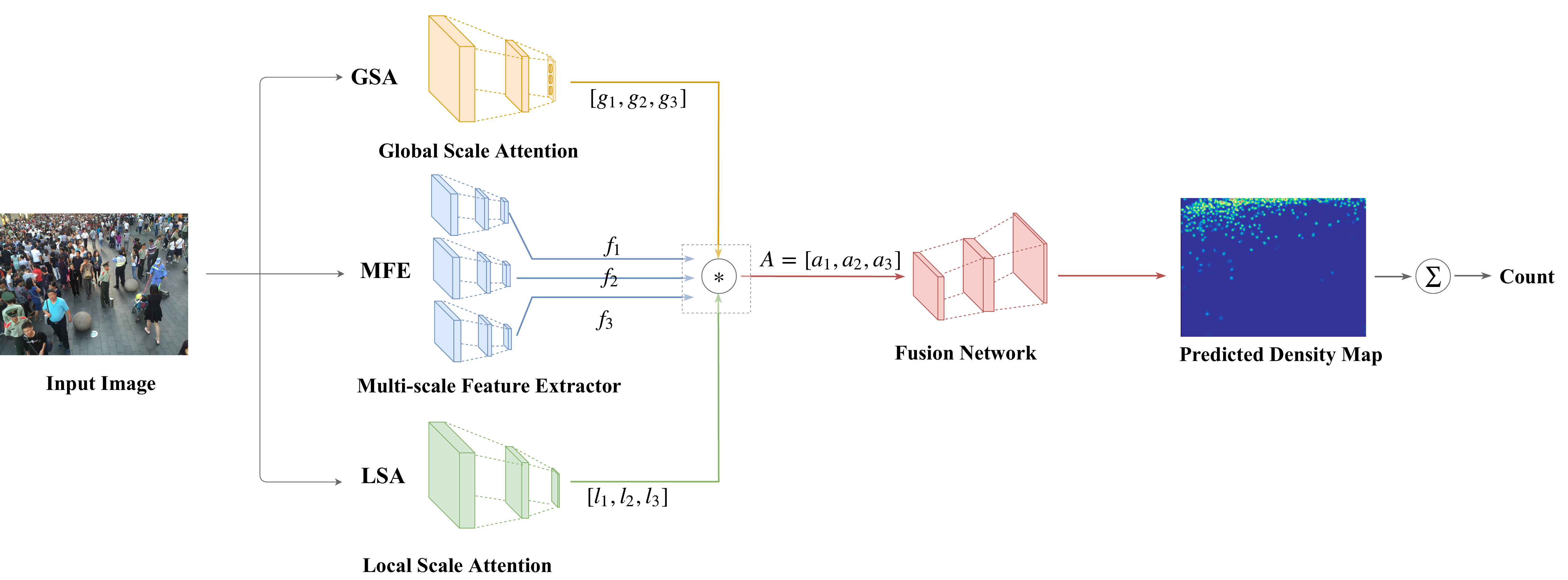}
\caption{Overview of the proposed method. The input image is passed simultaneously through three sub-networks: multi-scale feature extractor (MFE), global scale attentions (GSA), and local scale attentions (LSA). MFE extracts feature maps in three different scales. GSA and LSA produce three global scores and three pixel-wise local attention maps, respectively. Multi-scale feature maps are then weighted by the corresponding GSA and LSA outputs. The attention-weighted features are then used as the input to the fusion network to predict the density map. Finally, the crowd count can be obtained by summing over the entries in the predicted density map.}
\label{fig:overall}
\end{figure*}
\section{Our Approach}\label{sec:approach}
Crowd densities can vary dramatically both across different images and across different spatial locations within the same image. We propose to use both global and local attention weights to capture inter-image and intra-image variations of crowd density. This allows our model to adaptively use features at appropriate scales. Our proposed approach has several modules (see Fig~\ref{fig:overall}): multi-scale feature extractor (MFE), global scale attentions (GSA), local scale attention (LSA), and the fusion network (FN) for density estimation. Following, we describe each module in details.

\subsection{Multi-Scale Feature Extractor}\label{sec:feature}
\begin{figure*}[t]
\centering
\includegraphics[width=5.5in, height=2in]{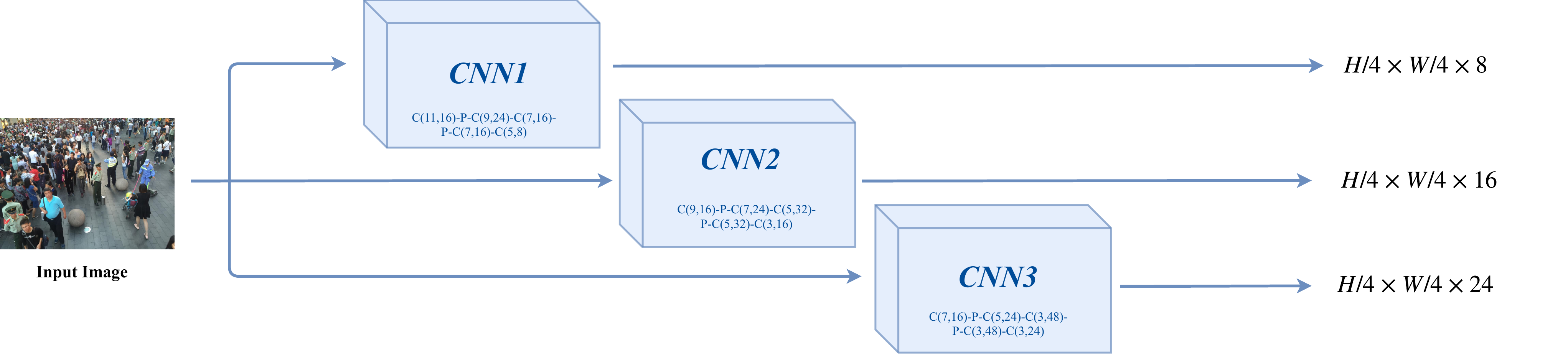}
\caption{Illustration of the multi-branch feature extractor (MFE) module. Here we use three branches associated with three different scales. Each branch uses different filter sizes to extract features at different scales. Using max-pooling, the output of each branch has a spatial dimension equal to one-fourth of the input image size. In this figure, we use $C(i,j)$ to denote a 2D convolution layer with filter sizes of $i\times i$ and $j$ output feature maps. We use $P$ to denote max-pooling.}
\label{fig:feature}
\end{figure*}

The goal of this module is to extract multi-scale feature maps from an input image. Inspired by the success of using multi-branch architectures in crowd-counting \cite{zhangY16_cvpr}, we use a similar multi-branch architecture to extract feature maps at different scales. The architecture (see Fig.~\ref{fig:feature}) has three branches associated with three different scales. It takes an image of arbitrary size as its input. Each branch then independently processes the input at its corresponding scale level. Each branch consists of multiple blocks of convolution layers with different filter sizes. By choosing different filter sizes in these branches, we can change the receptive field in each branch and capture the features at different scales. Each branch also has several max-pooling layers. We choose the filter sizes in max-pooling so that the output feature map in each branch has a spatial dimension equal to one-fourth of the input image dimension.

\subsection{Global Scale Attention}\label{sec:global}
\begin{figure*}[t]
\centering
\includegraphics[width=4in, height=2in]{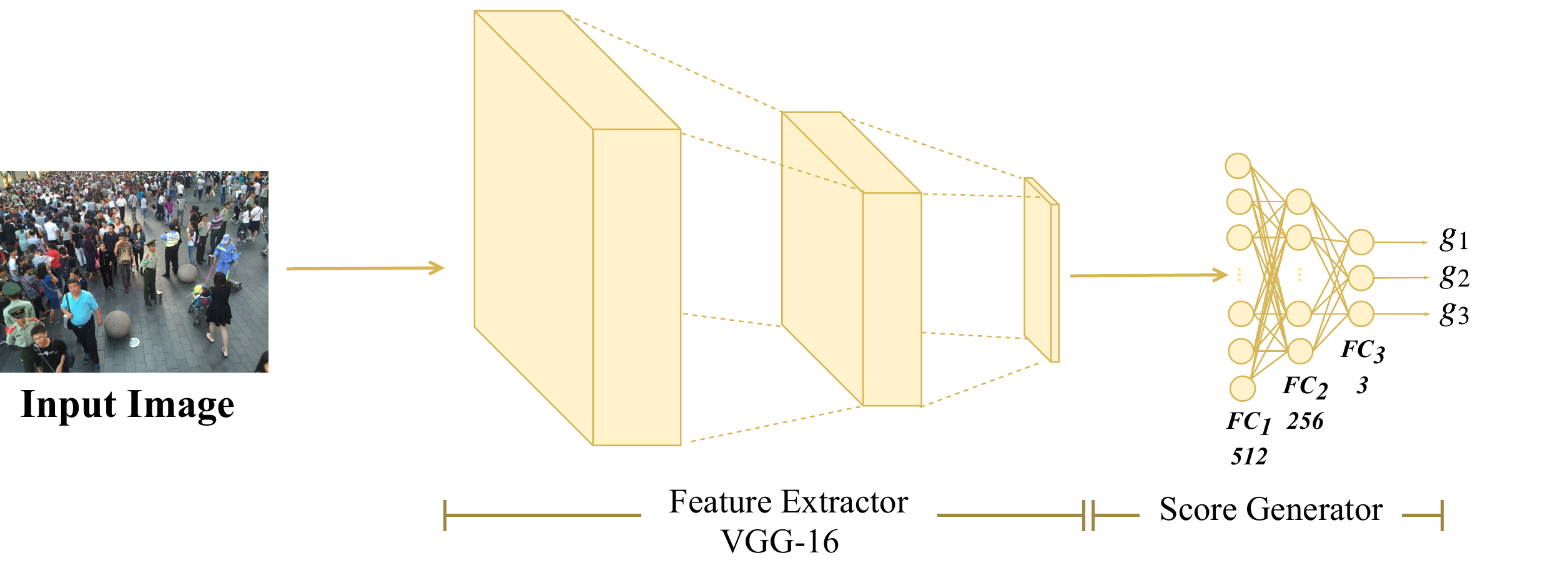}
\caption{Illustration of the global scale attention (GSA) module. Given an input image, this module generates three attention scores. Each score corresponds to one of the three pre-defined density labels: low-density, mid-density, and high-density. }
\label{fig:global}
\end{figure*}

Previous work \cite{sam17_cvpr,sindagi17_iccv,sindagi17_avss} has shown the benefit of leveraging global scale information in crowd counting problem. In our model, we use a global scale attention (GSA) module to capture global contextual information about how dense the image is. This module takes an input image and produces three attention scores. Each score corresponds to one of the three pre-defined density label level: low-density, mid-density, and high-density. The number of density levels is equal to the number of scales in the multi-scale feature extraction module (Sec. \ref{sec:feature}). The architecture of GSA is illustrated in Fig. \ref{fig:global}.

GSA outputs three scores $g_i$ ($i=1,2,3$) for each input image, representing the extent to which the input image belongs to each of the three density labels. A softmax layer is used at the end of the pipeline to normalize the scores (which can be interpreted as ``\textit{attentions}'') to sum to one. 
\subsection{Local Scale Attention}\label{sec:local}
The GSA module captures the overall density level of an image. But an image may have different density levels at different locations. The global density level may not sufficient to capture the fine-grained local contextual information at different locations in an image. Inspired by \cite{sindagi17_iccv}, we incorporate a \textit{local scale attention (LSA)} module to capture the local scale information at different locations in an image. LSA generates pixel-wise attention maps representing the scale information at different locations. Similar to the global attention module, here we also consider three different scale levels. Unlike GSA that produces three scalar scores, LSA will produce three pixel-wise attention maps. These attention maps have the same spatial dimensions as the corresponding multi-scale feature maps (Sec.~\ref{sec:feature}).

\begin{figure*}[t]
\centering
\includegraphics[width=3.7in, height=1in]{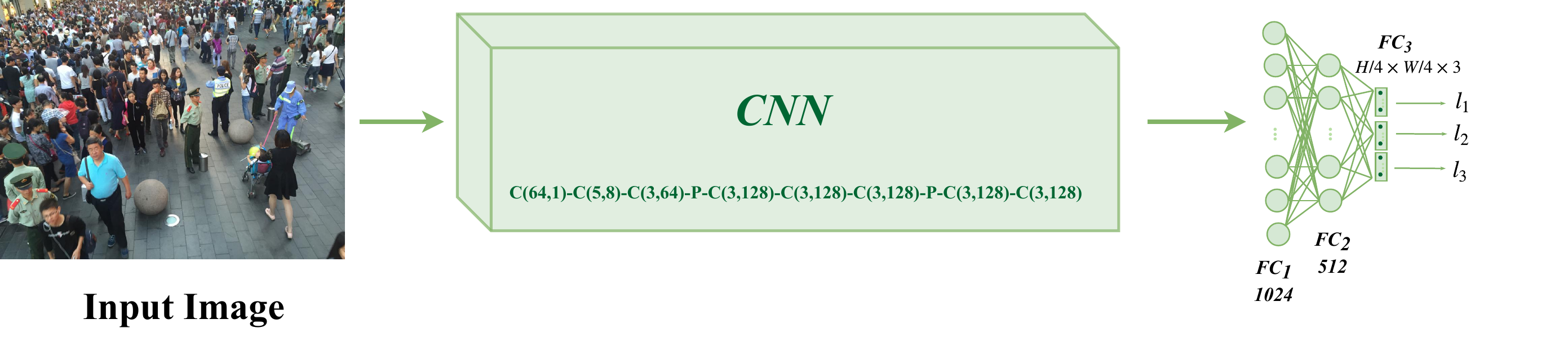}
\caption{Local Scale Attention Network. This module produces three feature maps of size ${H/4 \times W/4}$ each of which associated to a density scale. $C(i,j)$ denotes a 2D convolution with a filter size of $i\times i$ and with $j$ output feature maps. $P$ stands for max pooling.}
\label{fig:local}
\end{figure*}

The LSA module consists of eight convolution and two max-pooling layers, followed by three fully connected layers~(see Fig.~\ref{fig:local}). A sigmoid layer is placed at the end of the module to ensure the values in the attention maps are between 0 to 1. 
\subsection{Fusion Network}\label{sec:fusion}
The last component of the proposed method is the fusion network (FN) which produces the final density map for an input image. This module takes the extracted feature maps from the image which are re-weighted by the global and local attention scores. The output of this component is a predicted density map. The final crowd count can be obtained by summing over all entries of the estimated density map.

Let $f_i\in\mathbb{R}^{H_i\times W_i\times D_i}$ ($i=1,2,3$) denote the feature maps corresponding to the three different scales, where $H_i\times W_i$ is the spatial dimension and $D_i$ is the number of channels for the $i^{th}$ feature map. We use $g_i\in\mathbb{R}$ ($i=1,2,3$) and $l_i\in\mathbb{R}^{H_i\times W_i}$ ($i=1,2,3$) to denote the corresponding global and local attention scores, respectively. We can use the global and local attention scores to re-weight the feature maps. Let $f_{i}^{h,w,d}$ denote the $(h,w,d)$ entry of the feature map corresponding to the $i$-th scale. Similarly, let $l_{i}^{h,w}$ denote the $(h,w)$ entry of the corresponding local attention map $l_{i}$. We use $a_{i}\in\mathbb{R}^{H_i\times W_i\times D_i}$ to denote the attention-weighted feature map for the $i$-th scale. The $(h,w,d)$ entry $a_{i}^{h,w,d}$ of $a_{i}$ is calculated as follows:
\begin{equation}\label{eq:attention}
  a_{i}^{h,w,d}=g_{i}\cdot l_i^{h,w}\cdot f_i^{h,w,d}, \quad \textrm{where $i=1,2,3$}
\end{equation}
As shown in Fig. \ref{fig:feature} and described in Sec. \ref{sec:feature}, the extracted feature maps for all three scales are of size $H/4 \times W/4$ where $H\times W$ is the spatial dimension of the input image. However, the feature maps can have different depth (i.e. number of channels) among different scales. Here the depth dimension is set to 24, 16 and 8 for the low-density, mid-density, and high-density scales, respectively. 

The attention-weight feature maps for different scales are then concatenated together and fed into the fusion network (FN) module to produce the density map. The FN module consists of several convolution layers along with 2 de-convolution layers which resize the feature maps into size of $H/2\times W/2$ and eventually $H\times W$, where $H\times W$ is the spatial dimensions of the original input image. This output of this stage is a feature map of size $H\times W\times 16$. Finally, we apply a $1\times 1$ convolution to produce the density map ($DM$) of size $H\times W$ (i.e. of depth 1). To obtain the final crowd count from the 2-D density map $DM$, we simply sum over the entries in the density map as follows:
\begin{equation}\label{eq:sum}
Count = \sum_{j=1}^{H} \sum_{k=1}^{W} DM(j,k)
\end{equation}
where $DM(j,k)$ is the value of at the spatial position $(j,k)$ in the predicted density map.

\subsection{Loss Function}
In order to train the model parameters, we define an overall loss function $L_{final}$ consisting of three losses:
\begin{equation}\label{eq:final}
 L_{final}=L_{DM} + \lambda_g \cdot L_{GSA} + \lambda_l \cdot L_{LSA}
\end{equation}
In Eq.~\ref{eq:final}, $L_{DM}$ is a loss function defined on the predicted density map. This loss will encourage the model to predict density maps close to the ground-truth density maps on the training data. We also use two auxiliary losses $L_{GSA}$ and $L_{LSA}$. These two losses  will encourage the predicted global and local density scale attentions to be similar to the ground-truth global and local scales, respectively. The hyperparameters $\lambda_g$ and $\lambda_l$ are used to control the relative contributions of the two auxiliary losses. In the following, we provide details of these loss functions. To simply the notation, we focus on the definition of each loss function on one single training image. The final loss will be accumulated over all training images in the end.

Let $C\in\mathbb{R}^{H\times W}$ be the predicted density map on a training image and $C_{gt}\in\mathbb{R}^{H\times W}$ be the corresponding ground-truth density map. Here $H\times W$ is the spatial dimension of the input image. The loss $L_{DM}$ is defined as the square of the Frobenius norm between $C$ and $C_{gt}$, i.e. $L_{DM}=\frac{1}{2}||\mathrm{vec}(C)-\mathrm{vec}(C_{gt})||^2$ where $\mathrm{vec}(\cdot)$ concatenates entries of a matrix into a vector.

The auxiliary loss $L_{GSA}$ is used to encourage the predicted global scale attention scores to be close to the ground-truth global scale on a training image. We obtain the ground-truth global scale as follows. First, we find the maximum and minimum crowd count (denoted as $Count_{max}$ and $Count_{min}$, respectively) on the training data. We split the range $[Count_{min}, Count_{max}]$ into three bins of equal sizes. For a training image, we assign its ground-truth global scale $g_{gt}$ ($g_{gt}\in\{1,2,3\}$) according to the bin that the ground-truth crowd count falls into. Let $g\in\mathbb{R}^{3}$ be a vector of the global attention scores (corresponding to 3 different scales) on this training image. We can consider $g$ to be the score of classifying the image into one of the three global scales. We define $L_{GSA}$ using the standard cross-entropy loss as $L_{GSA}=CE(g,g_{gt})$ where $CE(\cdot)$ denotes the multi-class cross-entropy loss function.

The loss $L_{LSA}$ is used to to encourage the predicted local scale at each spatial location to be consistent with the ground-truth local scale on a training image. We generate the ground-truth local scale as follows. For a pixel location in a training image, we obtain a local crowd count at this location by summing over the ground-truth density map in the $64\times 64$ neighborhood of this location. We then find the minimum/maximum of local crowd count on training images. Similarly, we split the range to three bins and assign the ground-truth scale at a pixel location according to the bin that the local crowd count (over $64\times 64$ neighborhood) falls into. Let $l_{gt}\in\mathbb{R}^{H\times W}$ be a matrix of ground-truth local scales of a training image, where $H\times W$ denotes the spatial dimension. Each entry $l_{gt}$ ($l_{gt}\in \{1,2,3\}$) indicates the ground-truth local scale at the corresponding spatial location. Let $l\in\mathbb{R}^{H\times W\times 3}$ denote a tensor of predicted local scale attentions. We define $L_{LSA}$ as the sum of cross-entropy losses over all spatial locations, i.e. $L_{LSA}=\sum_{h=1}^{H}\sum_{w=1}^{W} CE(l[h,w,:], l_{gt}[h,w])$ using Matlab notations.

Empirically, we have found that the auxiliary losses help regularizing the model and improve the performance. In the experiments, we will provide ablation analysis on the impact of these auxiliary losses.

\section{Experiments}\label{sec:experiments}
We first introduce the datasets and our experimental setup. We then present experimental results on three benchmark datasets. Finally, we perform ablation studies to further analyze our proposed approach.

\subsection{Datasets and Setup}\label{sec:setup}
\noindent{\bf Datasets:} We evaluate our proposed method on three benchmark datasets: ShanghaiTech PartB~\cite{zhangY16_cvpr}, Mall dataset~\cite{mall} and UCF\_CC\_50 \cite{idrees13_cvpr}. Table~\ref{tab:dataset} shows various statistics of these datasets. The ShanghaiTech PartB dataset contains 716 where 400 images are for training and the other 316 ones for testing. The Mall dataset~\cite{mall} has 2000 frames captured from a shopping mall. The first 800 frames are used as training frames and the remaining 1200 frames are used for testing. The UCF\_CC\_50 dataset~\cite{idrees13_cvpr} contains a total of 50 images from web sources. Clearly, the limited number of images in these datasets raises the need for data augmentation in order to prepare the data for training a deep network. Consequently, we used data augmentation to address the issue. We followed the same data augmentation technique used in existing methods \cite{decidenet,idrees13_cvpr,zhangY16_cvpr,sam17_cvpr,sindagi17_iccv}.

\begin{table*}[!htbp]

\begin{center}
  \begin{tabular}{ | c  c  c  c  c  c  c |}
    \hline
    \textbf{Dataset} & \textbf{Resolution} & \textbf{Images} & \textbf{Max} & \textbf{Min} & \textbf{Avg} & \textbf{Total} \\ \hline
    
    ShanghaiTech PartB\cite{zhangY16_cvpr} & $1024\times768$ & 716 & 578 & 9 & 123.6 & 88,488 \\ 
    Mall\cite{mall} & $320\times640$ & 2000 & 53 & 13 & 33 &  62325 \\ 
    UCF\_CC\_50\cite{idrees13_cvpr} & varies & 50 & 4543 & 94 & 1279.5 & 63,974 \\ 
    \hline
  \end{tabular}
  \caption{Statistics of the three datasets used in the experiments. For each dataset, we show the following information: the image resolution, the number of images, the maximum/minimum number of people in an image, the average and total number of people annotated in the dataset.}
  \label{tab:dataset}  
\end{center}
\end{table*}

\noindent {\bf Evaluation Metric}: Following previous work \cite{zhangC15_cvpr,zhangY16_cvpr,sam17_cvpr,sindagi17_iccv,sindagi17_avss}, we use Mean Absolute Error (MAE) and Mean Square Error (MSE) as the evaluation metrics. Let $N$ be the number of test images, $\textrm{Count}_{gt}^{(n)}$ be the ground truth count and $\textrm{Count}^{(n)}$ be the predicted count for the $n$-th test image. These two evaluation metrics are defined as follows:
\begin{equation}
\text{MAE} =\frac{1}{N}\sum_{n=1}^{N} |Count^{(n)}-Count_{gt}^{(n)}|
\end{equation}
\begin{equation}
\text{MSE} =\sqrt[]{\frac{1}{N}\sum_{n=1}^{N} {|Count^{(n)}-Count_{gt}^{(n)}|}^2}
\end{equation}\\
\noindent {\bf Ground-truth Density Map}: On each dataset, head annotations (i.e. center of the head of a person) are provided as points. Following \cite{decidenet}, we generate the ground-truth density map from these point annotations by applying a Gaussian kernel normalized to have a sum of one.

\noindent {\bf Training Details:} We follow the data augmentation technique used in previous methods~\cite{decidenet,idrees13_cvpr,zhangY16_cvpr,sam17_cvpr,sindagi17_iccv}. Since there are lots of parameters in our model, directly learning all parameters from scratch is challenging. In our implementation, we use a two-phase training scheme to train our proposed model. During the first phase, we ignore the LSA module and only learn the parameters of the GSA, MFE, and FN modules. We assign each training image to one of three global scale classes (namely, low-density, mid-density, high-density) according to its ground-truth density map. We then learn the parameters of GSA, MFE, and FN modules by optimizing $L_{DM}+\lambda_{g}\cdot L_{GSA}$. During the second phase, we train all the modules (including LSA) together. The parameters of the GSA, MFE, and FN modules are initialized with the parameters obtained from the first phase. The parameters of the LSA module are initialized randomly from scratch.

\subsection{Experimental Results}\label{sec:results}
On the ShanghaiTech PartB dataset and the Mall dataset, we follow the standard training/testing split used in previous work~\cite{decidenet}. On the UCF\_CC\_50 dataset, we follow \cite{sindagi17_iccv} and perform 5-fold cross-validation. We apply the same data augmentation used in previous methods \cite{decidenet,idrees13_cvpr,zhangY16_cvpr,sam17_cvpr,sindagi17_iccv} on all datasets.

The experimental results on these three datasets are shown in Tables~\ref{shb_table}, Table~\ref{mall_table} and Table~\ref{ucf_table}, respectively. We also compare with existing state-of-the-art results on each of the datasets. On the ShanghaiTech PartB dataset and the Mall dataset, our proposed model significantly outperforms previous approaches in terms of both MAE and MSE. On the UCF\_CC\_50 dataset, our model outperforms previous approaches in terms of MAE. In terms of MSE, our model performs better than most previous approaches except for \cite{sindagi17_iccv}.

\begin{table}[t]
\centering
  \begin{tabular}{  |c  |c c|}
  \hline
            Method & 
            \multicolumn{2}{c|}{\shortstack{ShanghaiTech\\Part B}}\\
             & MAE & MSE \\
            \hline
            R-FCN$^\dag$ \cite{rfcn}& 52.35 & 70.12  \\
            Faster R-CNN$^\dag$ \cite{faster_rcnn} & 44.51 & 53.22 \\
            Cross-Scene \cite{zhangC15_cvpr} & 32.00 & 49.80\\
            MC-CNN \cite{zhangY16_cvpr} &26.40& 41.30\\
            Switching-CNN \cite{sam17_cvpr} & 21.60 & 33.40  \\
            CP-CNN \cite{sindagi17_iccv} & 20.1 & 30.1 \\
            FCN \cite{fcn} & 23.76 & 33.12\\
            DecideNet \cite{decidenet} & {20.75} & {29.42}  \\
            Ours & {\textbf{16.86}} & {\textbf{28.41}}  \\
            \hline
  \end{tabular}
  \caption{Comparison of the performance of different methods on the ShanghaiTech PartB dataset~\cite{zhangY16_cvpr}. $^\dag$These results are obtained from \cite{decidenet}.}\label{shb_table}

\end{table}

\begin{table}[t]
\begin{center}
  \begin{tabular}{  |c  |c c|}
  \hline
            Method & 
             
            \multicolumn{2}{c|}{Mall}\\
             & MAE & MSE  \\
            \hline
            DecideNet \cite{decidenet} & 1.52& 1.90 \\
            R-FCN$^\dag$ \cite{rfcn}&  6.02 & 5.46 \\
            Faster R-CNN$^\dag$ \cite{faster_rcnn} &  5.91 & 6.60 \\
            SquareChn Detector$^\dag$ \cite{square} & 20.55&439.10\\
            Count-Forest \cite{count_forest} & 4.40 &2.40 \\
            Exemplary-Density$^\dag$ \cite{exemplary_density} & 1.82 &2.74  \\
            Boosting-CNN \cite{walachwcnn} & 2.01& --\\
            Mo-CNN \cite{mocnn}& 2.75 & 13.40 \\
            Weighted\_VLAD \cite{weighted_vlad}&2.41&9.12 \\
            Ours & {\textbf{1.28}} & {\textbf{1.68}} \\
            \hline
  \end{tabular}
  \caption{Comparison of the performance of different methods on the Mall dataset~\cite{mall}. $^\dag$These results are obtained from \cite{decidenet}.}\label{mall_table}
\end{center}

\end{table}

\begin{table}[t]
\begin{center}
  \begin{tabular}{  |c  |c c|}
  \hline
            Method & 
            \multicolumn{2}{c|}{UCF\_CC\_50} \\
             & MAE & MSE  \\
            \hline
            Cross-Scene \cite{zhangC15_cvpr} &467.00 & 498.50 \\
            MC-CNN \cite{zhangY16_cvpr} & 377.60 & 509.10 \\
            Switching-CNN \cite{sam17_cvpr} &318.10 &439.20 \\
            CP-CNN \cite{sindagi17_iccv} & 295.80&{\bf320.90} \\
            Lempitsky et al. \cite{lempitsky10_nips}& 493.4&487.1  \\
            Idrees et al. \cite{idrees13_cvpr}& 419.5 & 541.6  \\
            Crowd-Net \cite{boominathan16_multimedia} &452.5&-- \\
            Hydra-2s \cite{rubio16_eccv}& 333.73 & 425.26\\
            Ours  &{\bf 271.60} & 391.00 \\
            \hline
  \end{tabular}
  \caption{Comparison of the performance of different methods on the UCF\_CC\_50 dataset~\cite{idrees13_cvpr}.}\label{ucf_table}
\end{center}
\end{table}

\begin{figure*}[t]
  \centering
  {\setlength\tabcolsep{1pt}
  \begin{tabular}{cccc}
   
    \includegraphics[width=1.4in,height=.84in]{qualitative_images/final_images/IMG_6.jpg}&
\includegraphics[width=1.4in,height=.84in]{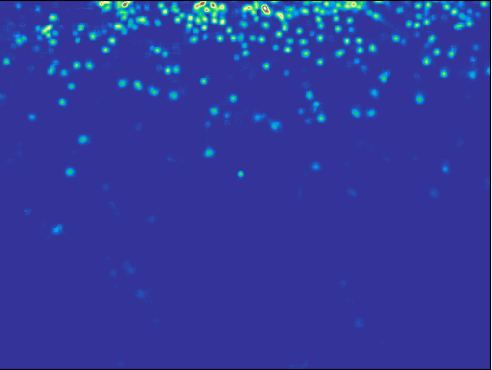}&    
    \includegraphics[width=1.4in,height=.84in]{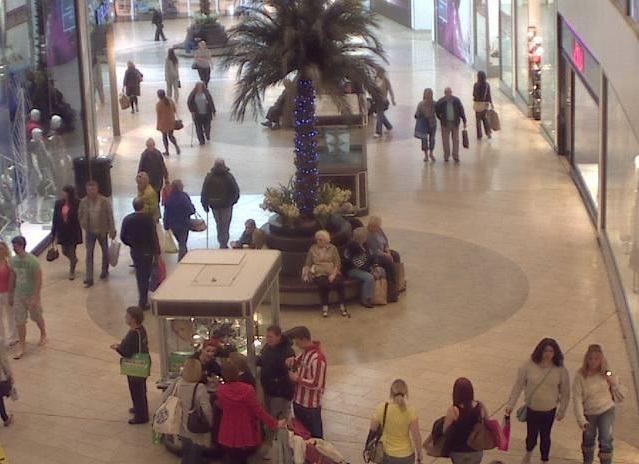}&    
    \includegraphics[width=1.4in,height=.84in]{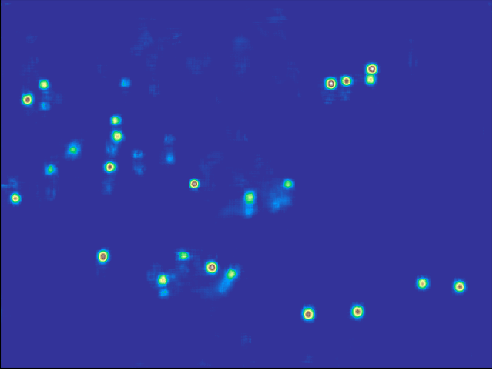}\\    
    \includegraphics[width=1.4in,height=.84in]{qualitative_images/final_images/IMG_12.jpg}&
    \includegraphics[width=1.4in,height=.84in]{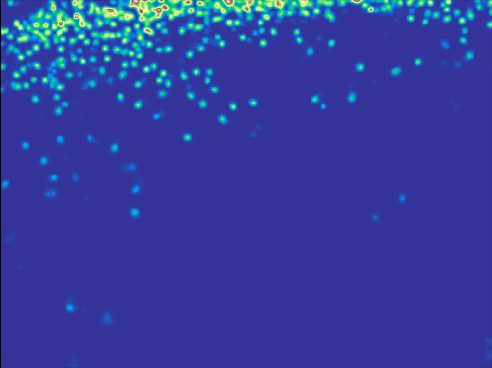}&    
    \includegraphics[width=1.4in,height=.84in]{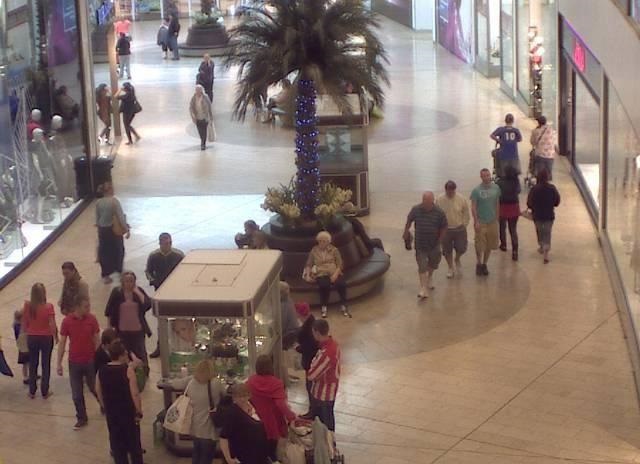}&
    \includegraphics[width=1.4in,height=.84in]{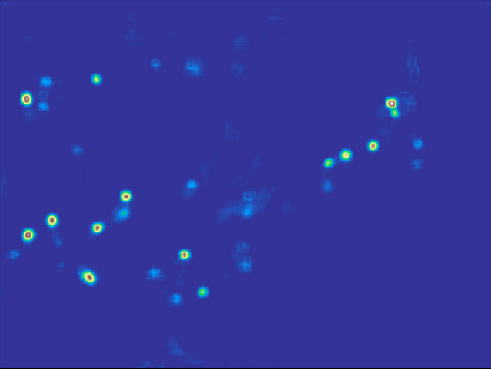}\\
    \includegraphics[width=1.4in,height=.84in]{qualitative_images/final_images/IMG_316.jpg}&
    \includegraphics[width=1.4in,height=.84in]{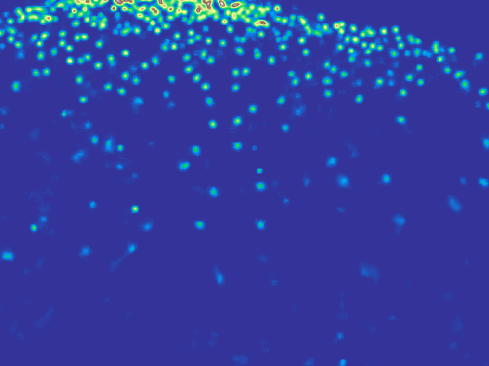}&
    \includegraphics[width=1.4in,height=.84in]{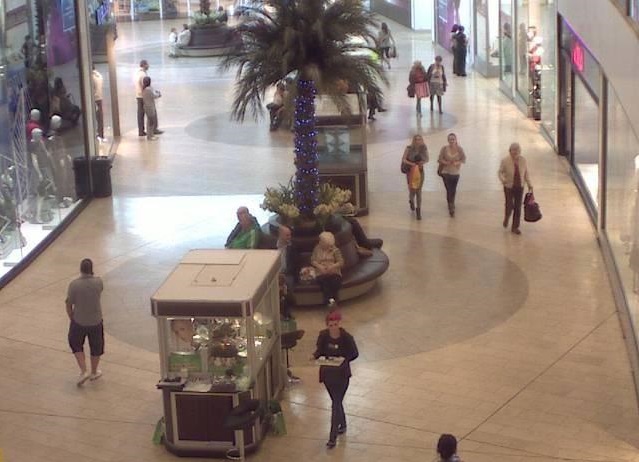}&
    \includegraphics[width=1.4in,height=.84in]{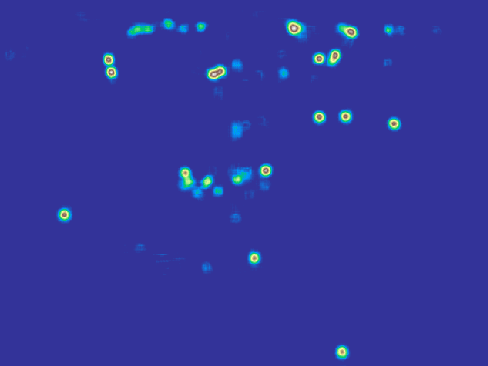}\\
    \includegraphics[width=1.4in,height=.84in]{qualitative_images/final_images/IMG_315.jpg}&
    \includegraphics[width=1.4in,height=.84in]{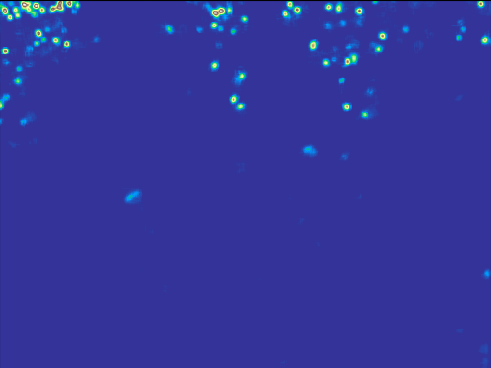}&
    \includegraphics[width=1.4in,height=.84in]{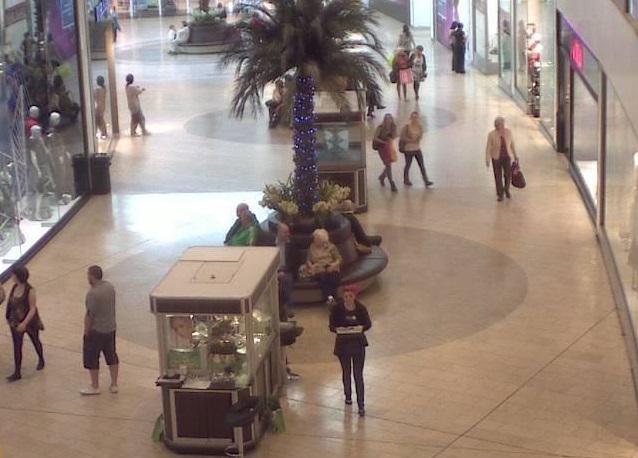}&
    \includegraphics[width=1.4in,height=.84in]{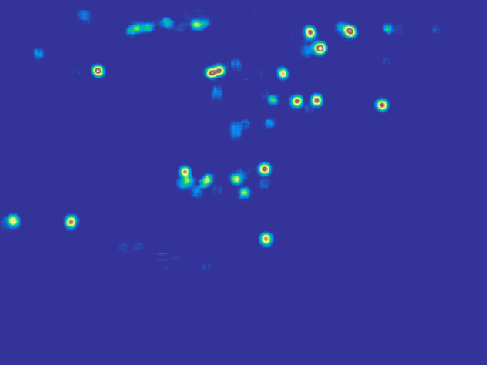}\\    
    \includegraphics[width=1.4in,height=.84in]{qualitative_images/final_images/IMG_314.jpg}&
    \includegraphics[width=1.4in,height=.84in]{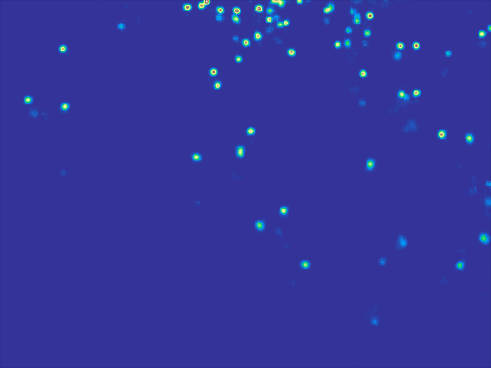}&
    \includegraphics[width=1.4in,height=.84in]{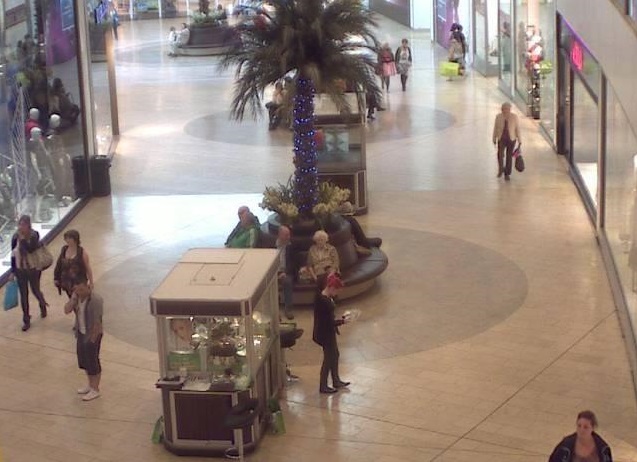}&
    \includegraphics[width=1.4in,height=.84in]{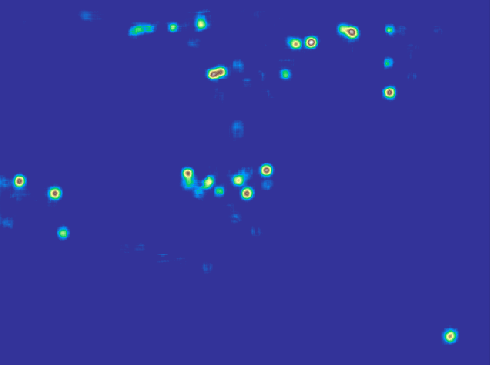}\\
    \includegraphics[width=1.4in,height=.84in]{qualitative_images/final_images/IMG_313.jpg}&
    \includegraphics[width=1.4in,height=.84in]{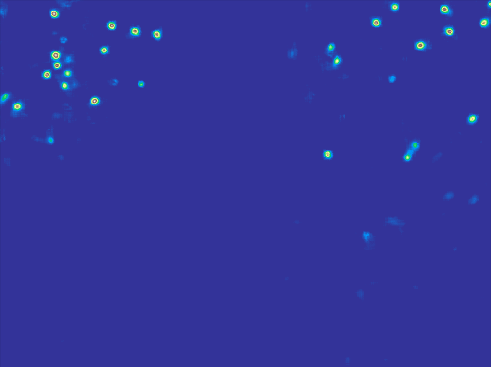}&    
    \includegraphics[width=1.4in,height=.84in]{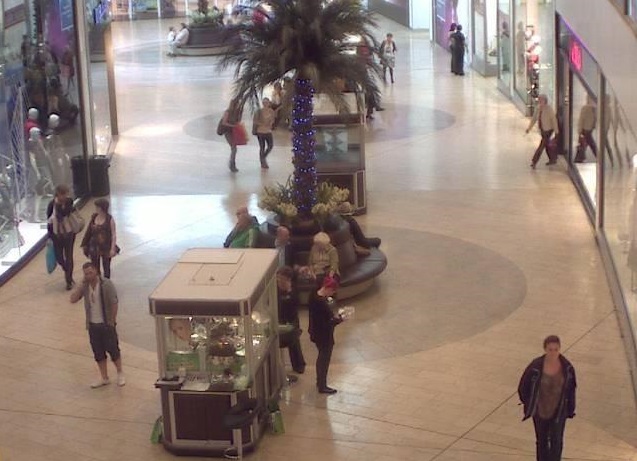}&
    \includegraphics[width=1.4in,height=.84in]{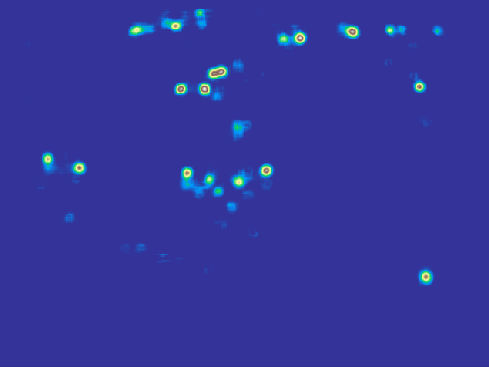}\\
    \includegraphics[width=1.4in,height=.84in]{qualitative_images/final_images/IMG_312.jpg}&
    \includegraphics[width=1.4in,height=.84in]{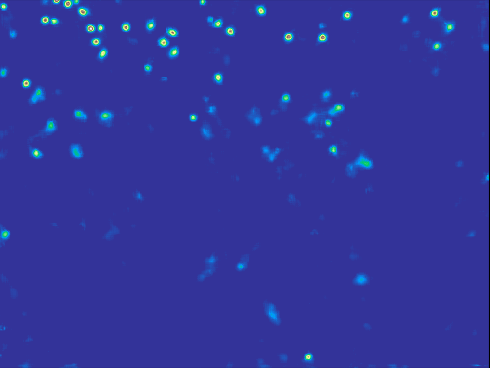}& 
    \includegraphics[width=1.4in,height=.84in]{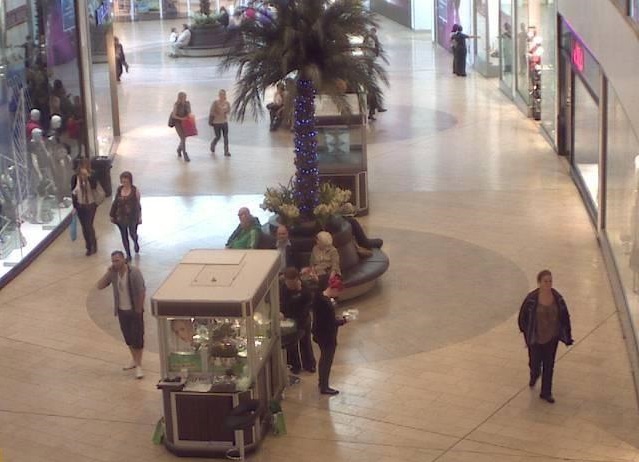}&
    \includegraphics[width=1.4in,height=.84in]{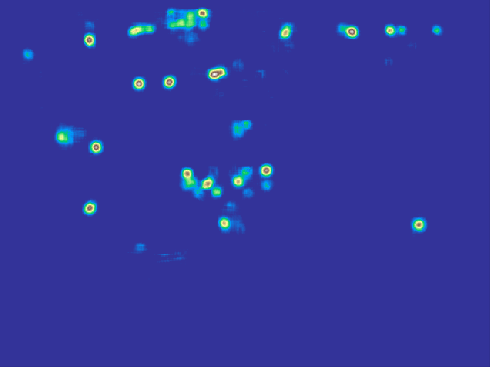}\\   
    \includegraphics[width=1.4in,height=.84in]{qualitative_images/final_images/IMG_311.jpg}&
    \includegraphics[width=1.4in,height=.84in]{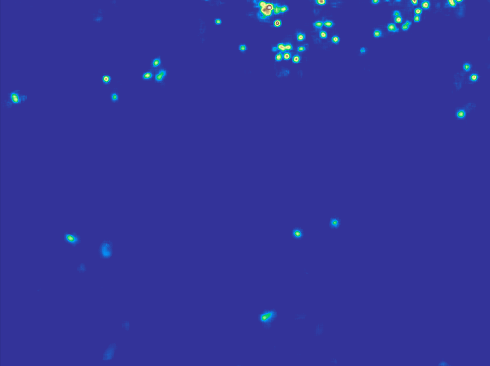}&
    \includegraphics[width=1.4in,height=.84in]{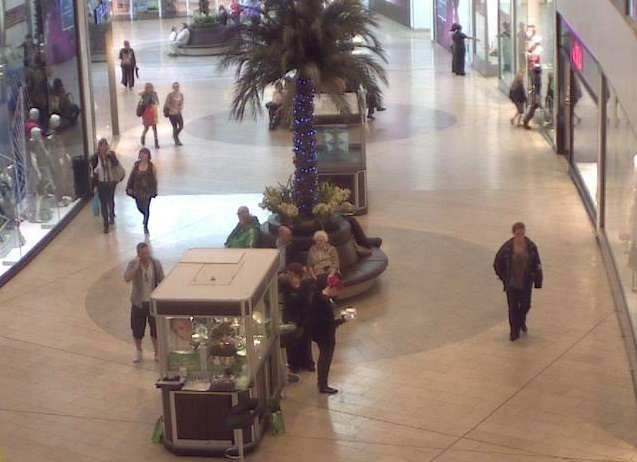}&
    \includegraphics[width=1.4in,height=.84in]{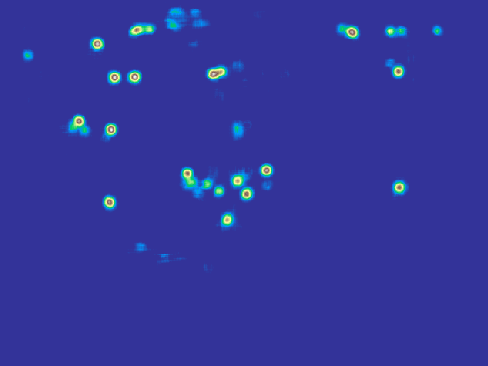}\\
    \includegraphics[width=1.4in,height=.84in]{qualitative_images/final_images/IMG_299.jpg}&
    \includegraphics[width=1.4in,height=.84in]{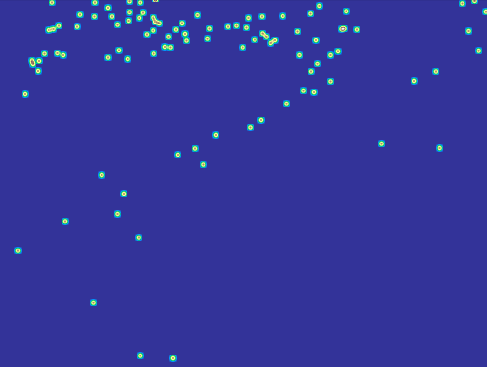}&
    \includegraphics[width=1.4in,height=.84in]{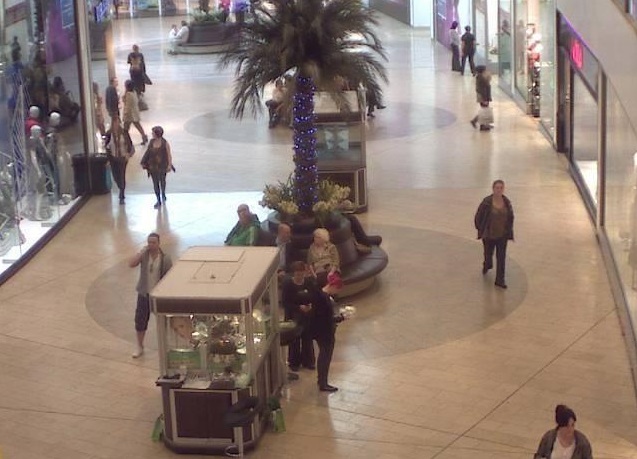}&
    \includegraphics[width=1.4in,height=.84in]{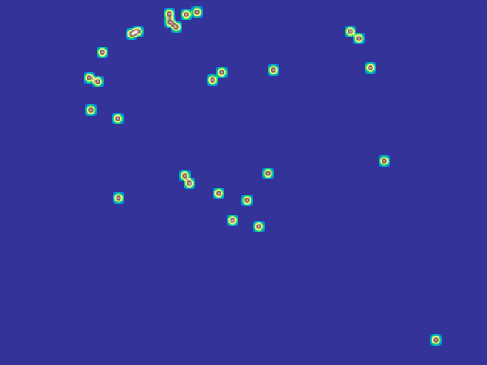}\\

  \end{tabular}
  }
  \caption{Qualitative examples of density maps. The first two columns are original input images from the Shanghaitech partB \cite{zhangY16_cvpr} dataset and their corresponding density maps respectively. The last two columns are original input images from the Mall \cite{mall} dataset and their corresponding density maps respectively.}
  \label{fig:localatt}
\end{figure*}
\subsection{Ablation Study}
We conduct ablation study on the ShanghaiTech PartB dataset to provide further analysis of relative contributions of various components of our approach.\\
\noindent {\bf Impact of attention modules}: First, we analyze the relative contributions of global and local attentions in our model. In Table~\ref{boost}, we show the results of removing the global or local attention modules (GSA or LSA) in our model. The base model in Table~\ref{boost} refers the architecture with only the multi-scale feature extractor and the fusion network, i.e. without any attention modules. From the results, we can see that both GSA and LSA contribute to the final performance. Using both GSA and LSA with the base model, we achieve the best performance.
\begin{table}[t]
\begin{center}
  \begin{tabular}{  |c  c  c|}
  \hline
            Methods
            & MAE & MSE\\
            \hline
            base model & 27.63 & 46.65\\
            base model+GSA & 17.0 & 30.60 \\
            base model+LSA & 18.07 & 31.87\\
            base model+GSA+LSA (this paper)  & \textbf{16.86} & \textbf{28.41} \\
            \hline

  \end{tabular}
  \caption{Effect of GSA and LSA modules on the ShanghaiTech PartB dataset. The ``base model'' only contains the multi-scale feature extractor and the fusion network, i.e. without any global or local attention modules. By adding GSA or LSA, we can achieve better performance. The best performance is obtained by using both GSA and LSA together with the base model.}\label{boost}
\end{center}
\end{table}
\\ 

\noindent {\bf Impact of auxiliary losses}: We also study the impact of using the auxiliary losses~(Eq.~\ref{eq:final}) as extra supervisions during training. In this analysis, we use our proposed model with all the modules. But during training, we use various combinations of the three losses: $L_{DM}$, $L_{GSA}$ and $L_{LSA}$. The results are shown in Table~\ref{cropsss}. We can see that both $L_{GSA}$ and $L_{LSA}$ help improving the performance of the learned model. By using all three losses, the best performance is achieved using all three losses during training.
\begin{table}[t]
\begin{center}
  \begin{tabular}{|ccc|}
  \hline
            Methods
             & MAE & MSE \\
            \hline
            $L_{DM}$ & 19.15 & 35.55\\
            $L_{DM}$ + $L_{LSA}$ & 17.02 & 31.49  \\
            $L_{DM}$ + $L_{GSA}$ & 17.33 & 32.33  \\
            $L_{DM}$ + $L_{LSA}$ + $L_{GSA}$ (this paper) &  {\bf 16.86} & {\bf 28.41}\\
            \hline
  \end{tabular}
  \caption{Effect of using auxiliary losses as extra supervisions on the ShanghaiTech PartB dataset. $L_{DM}$ is the model without any auxiliary losses. By adding $L_{LSA}$ (i.e. $L_{DM}$ + $L_{LSA}$) or $L_{GSA}$ (i.e. $L_{DM}$ + $L_{GSA}$ ), we can improve the performance. By using both $L_{LSA}$ and $L_{GSA}$ as auxiliary losses (i.e. $L_{DM}$ + $L_{LSA}$ + $L_{GSA}$), we achieve the best performance.}\label{cropsss}
\end{center}
\end{table}

\section{Conclusion}\label{sec:conclude}
We have presented scale-aware attention networks for crowd counting in images. The novelty of our work is that we use the attention mechanism to softly select the appropriate scales at both global and local levels. Compared with DecideNet~\cite{decidenet} that learns to switch between two different models (detection vs density estimation), our model is much simpler since it only has one model (density estimation). Our experimental results demonstrate that our proposed model outperforms other state-of-the-art approaches for crowd counting.

\\ \\
\textbf{Acknowledgment} {\small This work was supported by an NSERC Engage grant in 
collaboration with Sightline Innovation. We thank NVIDIA for 
donating some of the GPUs used in this work.}

{\small
\bibliographystyle{ieee}
\bibliography{wang}
}
\end{document}